\documentclass{article}

\usepackage[final,nonatbib]{neurips_2021}

\usepackage{comment}

\usepackage[utf8]{inputenc} 
\usepackage[T1]{fontenc}    
\usepackage{hyperref}       
\usepackage{url}            
\usepackage{booktabs}       
\usepackage{amsfonts}       
\usepackage{nicefrac}       
\usepackage{microtype}      
\usepackage{xcolor}         
\usepackage{endnotes}

\let\footnote=\endnote
\title{Ethics and Creativity in Computer Vision}

\author{%
  Negar Rostamzadeh\\
  Google Research\\
  Montreal\\
  \And
  Emily Denton\\
  Google Research\\
  New York\\
  \And
  Linda Petrini\\
  Google Research\\
  Montreal\\
}

\begin{document}

\maketitle

\begin{abstract}
This paper offers a retrospective of what we learnt from organizing the workshop \textit{Ethical Considerations in Creative applications of Computer Vision} at CVPR 2021 conference and, prior to that, a series of workshops on \textit{Computer Vision for Fashion, Art and Design} at ECCV 2018, ICCV 2019, and CVPR 2020. We hope this reflection will bring artists and machine learning researchers into conversation around the ethical and social dimensions of creative applications of computer vision.


\end{abstract}
\section{CVFAD: Computer Vision for Fashion, Art and Design}
In 2018, we organized, in conjuction with ECCV, the first workshop on Computer Vision for Fashion, Art and Design\footnote{https://sites.google.com/view/eccvfashion}. The workshop concentrated on generating, analyzing and processing visual content and invited the computer vision community to use generative model as a tool.
As part of this workshop, we organized the Fashion-Gen \cite{rostamzadeh2018fashion} challenge for language to visual fashion design. 
The workshop also included a Computer Vision Art Gallery\footnote{https://computervisionart.com/2018/}, organized by Luba Elliot and Xavier Snelgrove, to reflect the growing community of digital artists.
Overall, the workshop brought together researchers in computer vision, artists and professionals from creative domains to discuss open problems in the areas of computer vision for fashion and creative visual content generation. 

Our ICCV 2019\footnote{https://sites.google.com/view/cvcreative} and CVPR 2020\footnote{https://sites.google.com/view/cvcreative2020} workshops broadened the scope of focus to include economical and industrial applications of creative computer vision tools. We hosted two fashion oriented challenges, FashionIQ \cite{wu2021fashion} on multimodal fashion image retrieval \cite{wu2021fashion} by Wu et al and DeepFashion2 \cite{ge2019deepfashion2}\footnote{https://github.com/switchablenorms/DeepFashion2} on a variety of fashion and clothing tasks such as fashion landmark detection by Ge et al. In ICCV 2019, Adriana Kovashka\footnote{https://www.youtube.com/watch?v=Rqrpy1TU4z4} \cite{thomas2019predicting} brought up discussions on biases in creative advertisement content creation, and in political campaigns. We also had discussions during the panel discussions on potential harms that can be created by lack of representation in fashion industry and datasets. 

\section{EC3V: Ethical Considerations in Creative Applications of Computer Vision}
In 2021, we organized, in conjunction with CVPR, the first workshop on Ethical Considerations in Creative applications of Computer Vision (E3CV)
 \footnote{https://sites.google.com/view/ec3v-cvpr2021}. This workshop built upon 3 years of prior workshop organizing experience\,---\,4 creativity-oriented challenges, as well as a successful Creativity in AI workshop series at NeurIPS and multiple generative art symposiums at the intersection of machine learning research and fashion industry\,---\,and oriented our focus around the under-explored ethical dimensions of creative computer vision work. 
 
With this workshop, we brought together a team of computer vision researchers, artists, and sociotechnical researchers to address growing number of questions on broader impact of this research. At a high level, the workshop focused on (a) the recognition of creative computer vision technologies as a new form of art \footnote{https://www.youtube.com/watch?v=Jjv3m5oWICA}, (b) influence of these technologies on society and representations \cite{Malobola2021} and (c) the areas that requires greater attention and discussion, and can create potential harms \cite{prabhu2011taxonomy, srinivasan2021artsheets}. We also hosted the Computer Vision Art Gallery this year\footnote{https://computervisionart.com/} to bring more artists as one of the major stakeholders to the discussion. 

The computer vision art gallery showcased the work of 60+ artists addressing a range of topics and leveraging range of computer vision methods. For example, Jake Elwes's work `Zizi and Me'\footnote{zizi.ai}~\footnote{https://www.jakeelwes.com/project-zizi-and-me.html} showcased a double act between drag queen Me\footnote{https://www.instagram.com/methedragqueen/} and a deep fake clone of Me. In doing so, the work aims to both demystify AI and queer the process of AI development. Nouf Aljowaysir's project, `Salaf'\footnote{https://computervisionart.com/pieces2021/salaf/}, leveraged generative techniques to make visible the patterns of inclusion and exclusion operative in AI systems and express her personal frustrations with the Western colonial gaze so frequently embedded within the systems in use today. 

Many fruitful conversations came out of the workshop including the oral presentation \footnote{https://www.youtube.com/watch?v=tAea9kJRqcA} by Prabhu and Isiain \cite{prabhu2011taxonomy} critiquing our implementation of art gallery and submission process. Two main anchors that covered major conversations during our breakout sessions were around cultural appropriation, and ownership as well as issues integrated in training data.

\par
\textbf{Cultural appropriation vs inspiration}\\
Algorithmic techniques offer new routes for adopting and transferring aesthetic styles in ways that can be beautiful and creative and even shed light on cultures and art that individuals might not otherwise engage with. However, these same tools risk enabling new forms of cultural appropriation as they can make it even easier to extract from marginalized cultures without any accompanying investment in that culture, understanding of the significance of the artefact or aesthetics, or meaningful engagement with or say from the community. 

\par
Cultural appropriation \cite{Malobola2021}, distributed art and eventually generative art, all raise once again the question of intellectual property. If a traditional African pattern is re-contextualised in western fashion, is it fair to share the profit? If an artwork is the result of millions of people interacting with a website, should everyone get a fraction of the credit? The question becomes more complex when considering techniques based on deep learning, where a model is trained by exposure to thousands of images. These models can later be deployed by artists as aid to the creative process. How much credit goes to the ideator of the algorithm in this case? And how much to the creators of the content that was used to create the content? 
\par 
While ethical considerations for what concerns the work of a specific artist are starting to be discussed in depth, the aspect of broader cultural appropriation is still relatively unexplored. Fundamental challenges arise when trying to define ownership and copyrights in the context of Traditional Cultural Expression, where the intellectual contribution can't be attributed to a single individual, but results from, and often defines, the cultural evolution of specific groups of people. 

\par
\textbf{Generative art, training data: a source of inspiration or memorization?}\\
The ever-increasing ability of AI models to generate very realistic images introduces new challenges. 
AI-generated content can be carefully tailored to specific generations, creating (supposedly) new content that resembles the training data.
Considerations related to the training data are: biases inherently present in the data, memorization of training data, and insufficient transparency around some dataset creation processes \cite{srinivasan2021artsheets}. When using models trained on specific datasets to generate art, any bias present in the training data is unavoidably learned by the model and revealed in its generations \cite{srinivasan2021biases}. Many contemporary artists have begun to engage with ML researchers to find these biases. Some have done this for the visual aesthetics that the techniques begin to allow, others engage with them more critically in order to understand and reveal the algorithmic processes that are beginning to have great social and political power. 
These groups often notice the social dimension of the algorithms which can be overlooked by the computer science community.

\section{Conclusion}
There are several ethical considerations related specifically to creative domains. 
The introduction of AI within creative domains adds another layer of complexity to the growing discussions around social impacts of technologies. 
Some of social impact and considerations are driven by the generative capabilities of such models, some others from the issues concerning the training data. We hope to expand and discuss on these issues with the community of ML researchers, artists, and socio-technical researchers.

{\small
\bibliographystyle{ieee_fullname}
\bibliography{egbib}
}

\theendnotes


\appendix

\end{document}